%% file: arxiv.tex
\documentclass[final]{cvpr}
\usepackage{times}
\usepackage{epsfig}
\usepackage{graphicx}
\usepackage{amsmath}
\usepackage{amssymb}
\usepackage{url}
\usepackage{cite}
\usepackage{multirow}

\makeatletter\renewcommand\paragraph{\@startsection{paragraph}{4}{\z@}
  {.5em \@plus1ex \@minus.2ex}{-.5em}{\normalfont\normalsize\bfseries}}\makeatother

\def\eg{\textit{e.g.}}
\def\ie{\textit{i.e.}}

\usepackage[pagebackref=true,breaklinks=true,colorlinks,bookmarks=false]{hyperref}

\begin{document}
\title{Depth-conditioned Dynamic Message Propagation for \\Monocular 3D Object Detection}

\author
{ 
 Li Wang\textsuperscript{1}\thanks{The first three authors contributed equally to this work.}
\quad
 Liang Du\textsuperscript{1}$^*$
\quad
 Xiaoqing Ye\textsuperscript{2}$^*$
\quad
 Yanwei Fu\textsuperscript{1}
\quad
 Guodong Guo\textsuperscript{2}\\
\quad
 Xiangyang Xue\textsuperscript{1}
\quad
 Jianfeng Feng\textsuperscript{1}
\quad
 Li Zhang\textsuperscript{1}\thanks{
 Li Zhang (lizhangfd@fudan.edu.cn) is the corresponding author with School of Data Science, Fudan University.
 Li Wang and Xiangyang Xue are with School of Computer Science, Fudan University.
 Yanwei Fu is with the School of Data Science, MOE Frontiers Center for
Brain Science, and Shanghai Key Lab of Intelligent Information Processing, Fudan University.
 Liang Du and Jianfeng Feng are with the Institute of Science and Technology for Brain-Inspired Intelligence, Fudan University.
 }
 \\ 
 \textsuperscript{1}Fudan University
\quad
 \textsuperscript{2}Baidu Inc.
}

\maketitle

\input{file/0-abstract}
\input{file/1-intro}
\input{file/2-related}
\input{file/3-method}

\input{file/4-experiment}

\input{file/5-conclusion}

{\small
\bibliographystyle{ieee_fullname}
\bibliography{egbib}
}

\clearpage

\appendix
\section*{Appendix}

\input{file/6-appendix}

\end{document}

%% file: file/0-abstract.tex
\begin{abstract}
The objective of this paper is to learn context- and depth-aware feature representation to solve the problem of monocular 3D object detection.
We make following contributions:
(i) rather than appealing to the complicated pseudo-LiDAR based approach, we propose a depth-conditioned dynamic message propagation (DDMP) network to effectively integrate the multi-scale depth information with the image context;
(ii) this is achieved by first adaptively sampling context-aware nodes in the image context and then dynamically predicting hybrid depth-dependent filter weights and affinity matrices for propagating information;
(iii) by augmenting a center-aware depth encoding (CDE) task, our method successfully alleviates the inaccurate depth prior;
(iv) we thoroughly demonstrate the effectiveness of our proposed approach and show state-of-the-art results among the monocular-based approaches on the KITTI benchmark dataset.
Particularly, we rank $1^{st}$ in the highly competitive KITTI monocular 3D object detection track on the submission day (November 16th, 2020).
Code and models are released at \url{https://github.com/fudan-zvg/DDMP}
\end{abstract}

%% file: file/1-intro.tex
\section{Introduction}

Object detection is a fundamental problem in computer vision. Although promising progress in 2D object detection has been made \cite{ren2015faster, he2017mask, wang2017evolving, liu2016ssd, tian2019fcos} with convolutional neural networks (CNNs) in recent years, 3D object detection that perceives 3D object location, physical dimension, and orientation, still remains challenging and critical in applications such as autonomous driving \cite{geiger2012we, Du_2019_ICCV}, robotic grasp and navigation \cite{valassakis2020crossing, du20203dcfs}, and Mixed Reality (MR) \cite{mx_2019}.

\begin{figure}
 \centering
 \includegraphics[width=7.5cm]{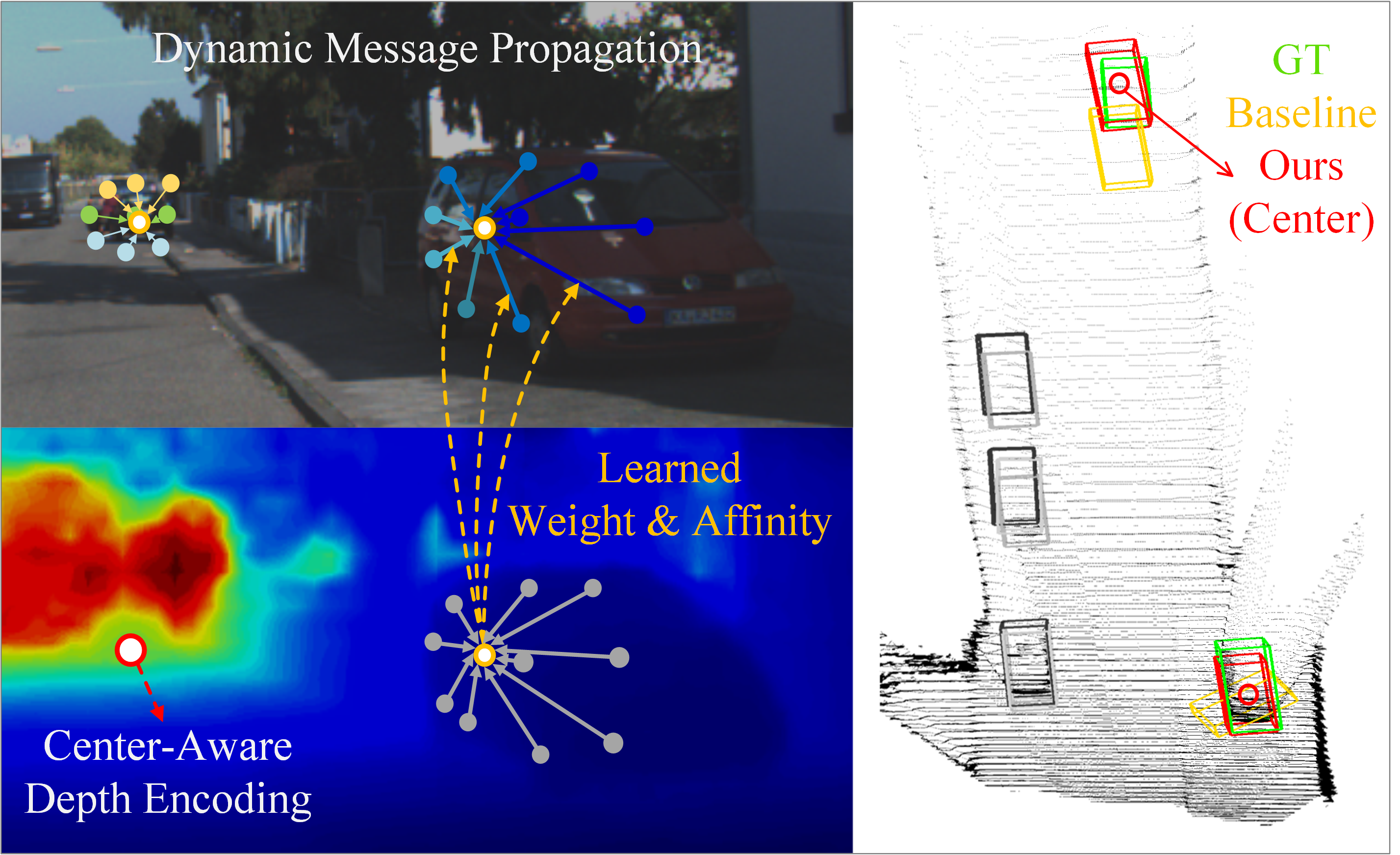}
 \caption{ \textbf{Left}: DDMP adaptively samples context-aware nodes (top) in the image context and dynamically predicting hybrid depth-dependent filter weights and affinity matrices (bottom) for propagating information.
 \textbf{Right}: the improvement of DDMP-3D (red) over the baseline (yellow) via center-aware depth encoding. }
 \label{fig_intro}
 \vspace{-0.15in}
\end{figure}

% introduce image-only
LiDAR point cloud based methods
\cite{ku2018joint, zhou2018voxelnet,yan2018second,shi2019pointrcnn, Du_2020_CVPR} have excelled in 3D object detection and achieve superior performance,
however, they still depend on the expensive LiDAR sensors and sparse data representation to make them scalable.
Cheaper alternative such as perceiving RGB images that are captured by the monocular camera is drawing increasing attention. Some image-only based methods \cite{mousavian20173d,brazil2019m3d,liu2019deep,chen2020monopair,brazil2020kinematic} aim to explore the 2D-3D geometric consistency for recovering reasonable 3D detection. Nevertheless, the performance is still far from satisfactory.
This is because
(i) scale variance caused by the perspective projection.
The monocular views at far and near distance cause significant changes in object scales.
It is difficult for conventional CNNs (\eg, 2D conv) to process objects of different scales.
(ii) lack of depth cues for the CNNs to capture the depth-aware feature for 3D reasoning.

% introduce pseudo-LiDAR
Recent efforts have been made to pursue pseudo-LiDAR based approaches~\cite{wang2019pseudo, ma2019accurate, weng2019monocular, you2019pseudo}.
The pseudo-LiDAR approaches first transform depth maps estimated from 2D images to point cloud data representations 
and then adopt existing LiDAR-based 3D detectors for prediction. %\cite{weng2019monocular, you2019pseudo}. 
Although improved performances on monocular 3D detection have been observed, they still suffer from the gap between inaccurate estimated depth and real-world depth.
Additionally, LiDAR-based approaches only employ 3D spatial information to generate LiDAR-resembled point cloud but discard the semantic information from RGB images, which is a vital clue to perceive and distinguish objects.

% introduce D4
Another line of research~\cite{ding2019learning, shi2020distance,ouyang2020dynamic} focuses on the perspective of image and depth fusion learning.
Arguably, with the depth assisted, the model can learn depth-aware feature representation, and both semantic and structure knowledge can be integrated for 3D reasoning.
Specifically, $\rm D^4LCN$~\cite{ding2019learning} proposes to generate dynamic-depthwise-dilated kernels from depth features to integrate with image context.
However, two issues still remain unsolved,
(i) Its empirical design can not guarantee the discriminative power of the model and the local dilated convolution is not be able to fully capture the object context in the condition of perspective projection and occlusion.
(ii) Its 3D detection performance heavily relies on the precision of estimated depth maps.
The model has no clue to resolve the inferior 3D localization caused by the inaccurate depth prior.

To this end, for the first time we present a graph-based formulation, a novel depth-conditioned dynamic message propagation model (DDMP-3D), to effectively learn depth-aware feature representations for monocular 3D object detection.
As shown in figure~\ref{fig_intro}, specifically, considering each feature pixel as a node within a graph, we first dynamically sample the neighbourhoods of a node from the feature graph.
This operation allows the network to gather the object contexts efficiently by adaptively selecting a subset of the most relevant nodes in the graph.
For the sampled nodes, we further predict filter weights and affinity matrices dependent on the aligned depth feature to propagate information through the sampled nodes.
Moreover, multi-scale depth features are explored over the propagation process, hybrid filter weights and affinity matrices are learned to adapt various scales of objects.

Additionally, to resolve the challenge of inaccurate depth prior, 
a center-aware depth encoding (CDE) is augmented as an auxiliary task append at the depth branch.
It performs 3D object center regression task which explicitly guides the intermediate features of the depth branch to be \textit{instance-aware} and further improves the localization of objects.

%% file: file/2-related.tex
\section{Related work}

\begin{figure*}[t]
 \centering
  \includegraphics[width=0.9\linewidth]{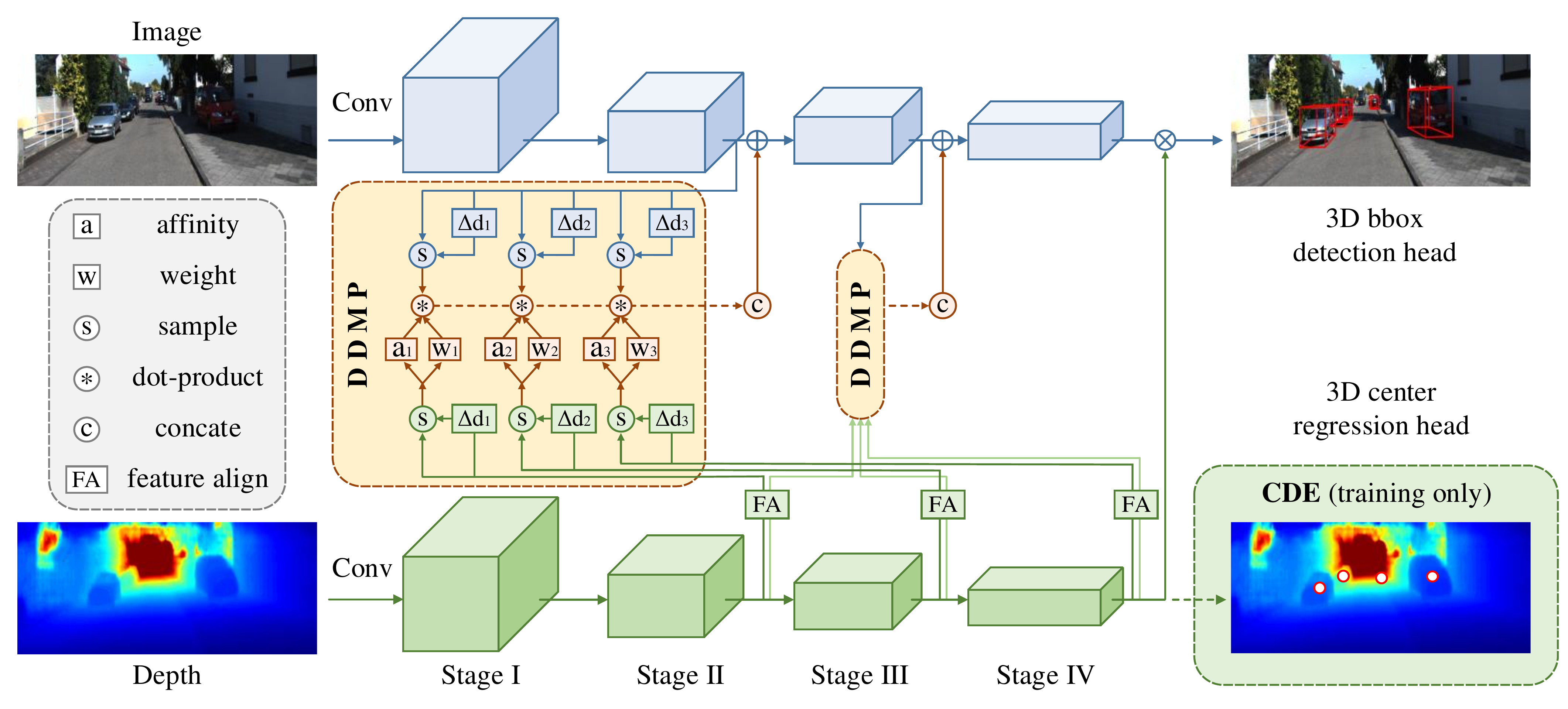}
 \caption{Schematic illustration of our proposed DDMP-3D.
 Two branches are involved including 3D detection branch (colored in blue) and depth feature extraction branch (colored in green). 
 The DDMP modules in yellow color reveal the depth-conditioned dynamic message propagation.
 It dynamically samples context-aware nodes in the upper image branch and predicts the hybrid filter weights and affinities based on multi-scale depth features from the bottom branch for message propagation.
 CDE is the auxiliary task for joint-optimization training and is discarded during inference.
}
 \label{fig_arch}
\end{figure*}

\noindent \textbf{Image-only 3D detection}
Due to the absence of accurate depth information, the monocular 3D detection task remains challenging.
Several works rely only on RGB information and geometry consistency to predict 3D boxes.
For instance, Deep3DBox \cite{mousavian20173d} utilizes bins classification to solve the orientation regression and enforces the 2D-3D box constraint to recover 3D locations and shape. M3D-RPN \cite{brazil2019m3d} leverages the geometric relationship between 2D and 3D perspectives by sharing the prior anchors and classification targets. For a better understanding of the spatial relationship between original 3D bounding boxes and the object, FQNet \cite{liu2019deep} infers the 3D IoU between 3D proposals and the object by drawing the projection results on the image plane to bring additional information. However, the orientation predictions can affect the following location prediction greatly, leading to a coupled 3D parameter regression. MonoPair \cite{chen2020monopair} captures spatial relationships between paired objects to improve accuracy on occluded objects, inspired by the key point-based CenterNet\cite{zhou2019objects}.

\noindent \textbf{Depth-assisted 3D detection} 
An alternative solution to improve monocular detection performance is to utilize depth information. The pioneering work pseudo-LiDAR \cite{wang2019pseudo, ma2019accurate} imitates the process of LiDAR-based (Point Cloud) 3D detection approaches by first estimating depth maps for input images from off-the-shelf methods such as DORN \cite{fu2018deep}, then transforming it into 3D space and adopting LiDAR-based 3D detection approaches.
Mono3D-PLiDAR \cite{weng2019monocular} points out that the noise in pseudo-LiDAR data is a bottleneck to improve performance and adopts the instance mask instead of the bounding box for frustum lifting.
With the help of the stereo network, more accurate depth estimation is achieved to aid the pseudo-LiDAR methods in \cite{you2019pseudo}.
Different to pseudo-LiDAR approaches that rely heavily on the accuracy of estimated depth, $\rm D^4LCN$ \cite{ding2019learning} carefully designs a local convolutional network, where the depth map was regarded as guidance to learn local dynamic depthwise-dilated kernels for images. 
However, we argue that the local dilated convolution is not able to fully capture the object context in the condition of perspective projection and occlusion, and the model has no mechanism to resolve the inferior 3D localization caused by the inaccurate depth prior.

In addition to the above-mentioned approaches, multi-task or leveraging auxiliary knowledge is exploited to improve the 3D detection performance. 
For instance, AM3D \cite{ma2019accurate} designs two modules for background segmentation and RGB information aggregation respectively in order to enhance the 3D box estimation task.

\noindent \textbf{Graph neural network}
A number of methods have been explored to model context for computer vision tasks, such as dilated convolution \cite{yu2015multi} and deformable convolution \cite{dai2017deformable}. 
Graph neural networks \cite{kipf2016semi,velivckovic2018graph,zhang2020dynamic}, on the other hand, propagate information along graph-structured input data. These networks are superior to both fronts on capturing object context.
In this paper, we present a graph-based formulation for depth-aware feature representation learning, with the goal of solving monocular 3D object detection.

%% file: file/3-method.tex
\section{Methodology}

We first briefly introduce the graph message passing formulation.
Then the overall framework and each critical component are explained in detail, including depth-conditioned dynamic message propagation (DDMP) and center-aware depth feature encoding (CDE).
Finally, we describe the instantiation of our model and loss functions.

\subsection{Graph message passing}
 The message passing mechanism constructs a feature graph $\mathcal{G=\{V,E},A\}$, where $\mathcal{V}$ stands for nodes, $\mathcal{E}$ for edges and $A$ for adjacency matrices. 
%  Be specific, 
 Each node in graph are represented as the latent feature vector $\mathbf{h}_i$, i.e., $\mathcal{V} = \{\mathbf{h_i}\}_{1}^{N}$, where $N$ is total number of nodes. And $A \in R^{N \times N}$ is a binary or learnable matrix with self-loops describing the connections between nodes. Network aims to refine latent feature vectors $\mathbf{h}_i$ by extracting hidden structured information among the feature vectors at different node locations. The common message passing phase takes $\mathit{T}$ iterations and is composed of a message calculation step $M^t$ and a message updating step $U^t$. Given a latent feature vector $\mathbf{h}_i^{(t)}$ at iteration $\mathit{t}$, it samples $\mathit{K}$ locally connected node field $v_{j} \subset \mathcal{V}, v_{j} \in R^{K \times C}$, where $C$ is vector dimension and $K \ll N$. Thus the message calculation step for node $i$ is:

\begin{equation}
\begin{aligned}
	\mathbf{m}_i^{t+1} & = M^t(A_{i,j} \{\mathbf{h}_1^{(t)},...,\mathbf{h}_K^{(t)}\},\mathbf{w}_j)\\
	& = \sum_{j \in \mathcal{N}(i)} A_{i,j} \mathbf{h}_j^{(t)} \mathbf{w}_j
	\label{messagecal}
\end{aligned}
\end{equation}
where $A_{i,j}$ is the connection relationship between latent nodes $\mathbf{h}_i^{(t)}$ and $\mathbf{h}_j^{(t)}$, $\mathcal{N}(i)$ contains the $\mathit{K}$ number of sampled nodes for $v_i$, and $\mathbf{w}_j \in R^{C \times C}$ is a transformation matrix for message calculation on the hidden node $\mathbf{h}_j^{(t)}$. Then message updating step $U^t$ updates the node $\mathbf{h}_i^{(t)}$ with a linear combination of the calculated message and the original node status:
\begin{equation}
\begin{aligned}
	\mathbf{h}_i^{t+1} = U^t(\mathbf{h}_i^{t}, \mathbf{m}_i^{t+1}) = \sigma(\mathbf{h}_i^{t}+ {\alpha}_i^m\mathbf{m}_i^{t+1})
	\label{messageupd}
\end{aligned}
\end{equation}
where ${\alpha}_i^m$ is a learnable parameter for scaling the message, and the operation $\sigma(\cdot)$ is a non-linearity function \textit{e.g.,} ReLU. By updating nodes T times, the network finally obtained refined features via message passing on each nodes.

\subsection{Framework overview}
We give an overview of the proposed DDMP-3D as in Figure~\ref{fig_arch}. It consists of two branches, one for 3D regression (the upper branch colored in blue) while the other for depth feature extraction (the bottom branch colored in green).
The RGB images are initially fed into the upper branch for feature extraction while corresponding depth maps estimated via off-the-shelf depth estimator are sent into the depth branch for extracting depth-aware features. 
Specifically, we first adaptively sample context-aware nodes in the image graph and then dynamically predict hybrid depth-dependent filter weights and affinity matrices. 
After obtaining the context- and depth- aware features from these two branches, we integrate them in a graph message propagation pattern via the DDMP module.
Common 3D heads for 3D center, dimension, orientation regression are followed to achieve final 3D object boxes. Moreover, an auxiliary center regression task is performed during the training process to implicitly guide the depth sub-network to learn center-aware depth features for better object localization.

\begin{figure}
 \centering
 \includegraphics[width=1\linewidth]{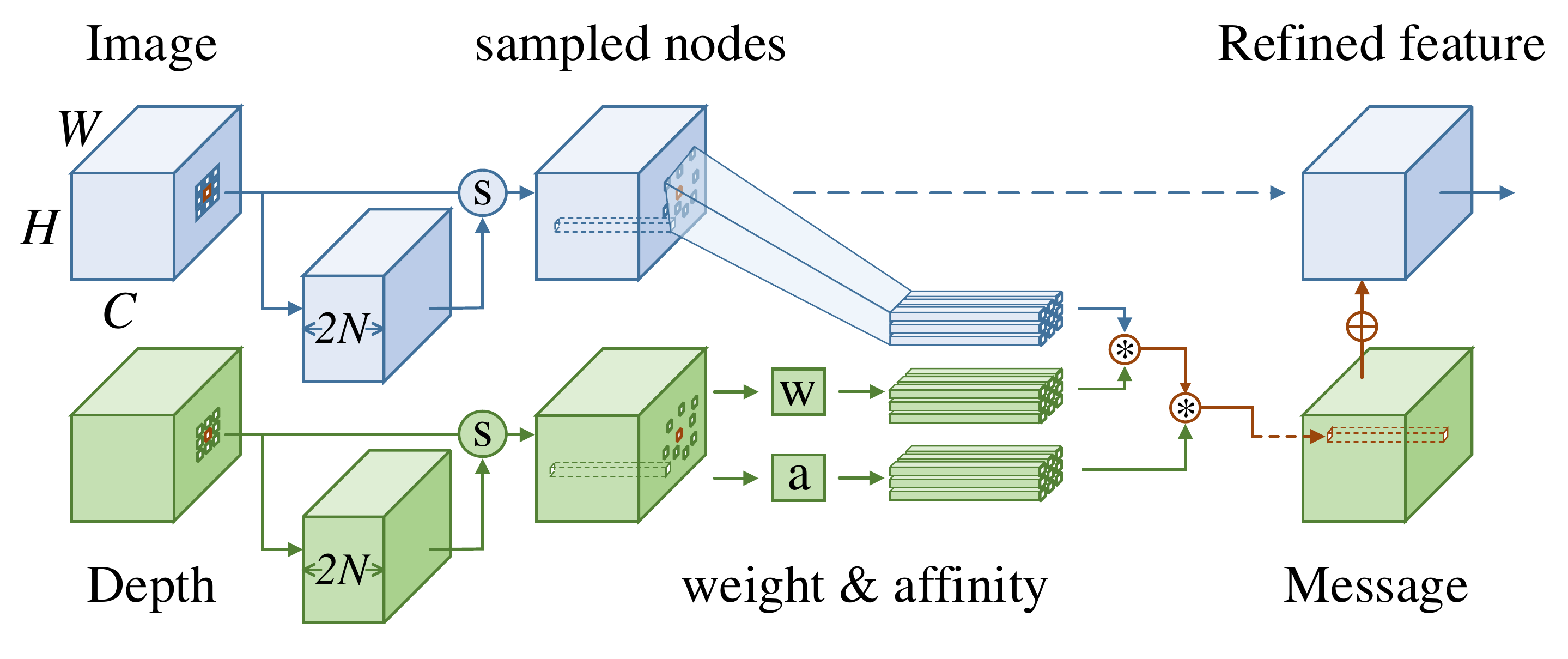}
 \caption{Illustration of DDMP module in a single scale pattern. Dynamic nodes are first sampled from the image and depth feature graph, for these sampled nodes, the filter weights and affinity matrices are learned from depth features to propagate the depth-conditioned message.}
 \label{fig_dgmi}
\end{figure}

\subsection{Depth-conditioned dynamic message propagation (DDMP)} 
We first introduce the process of our 
DDMP.
The input feature maps are regarded as a graph, in which each pixel of the feature map is a node vector $v_i \in R^{C}$ and all the pixels make up the set $\mathcal{V} = \{v_i\}_{1}^{N}$, where $C$ is the channel numbers of the input feature maps and $N$ is the total number of pixels, and its status at time $t$ is denoted as latent node $\mathbf{h}_i^{(t)}$.
Hence, depth-conditioned message propagation is employed on feature nodes to deliver the context- and depth-aware information. 
In particular, 
it includes two steps: 
1) Image feature node sampling, which selects a subset of relevant object nodes in the graph; 
2) For these sampled nodes, the hybrid filter weights and affinity matrices are learned from multi-scale depth feature to enrich the propagated message.

Figure~\ref{fig_dgmi} shows the depth-conditioned dynamic message propagation process. We first sample the context-aware nodes in the image graph. Due to the diverse scales and occlusion of targets caused by perspective projection, we follow the strategies in \cite{zhang2020dynamic, dai2017deformable} to explore dynamic sampling during the message propagation process. For each node $v_i$, 
the sampling number $K$ determines the receptive field of $v_i$.
To adaptively sample relevant nodes for $v_i$ while taking different feature distribution into consideration, the network learns to walk around so as to choose the most effective nodes among the uniformly distributed neighboring nodes. 
We denote $\Delta\mathbf{d}_{i,j} \in R^{D}$ as predicted walk for a uniformly sampled relevant node $v_{i,j}$ with $j \in \mathcal{N}(i)$, where $\mathcal{N}(i)$ contains $K$ number of sampled nodes for $v_{i}$ and $D=2$ is the space dimension along height and width. 
Then the node walk can be described as a matrix transformation:
\begin{equation}
\begin{aligned}
	\Delta\mathbf{d}_{i,j} = \mathbf{W}_{i,j} \mathbf{h}_i + \mathbf{b}_{i,j}
	\label{deltad}
\end{aligned}
\end{equation}
where $\mathbf{W}_{i,j}$ and $\mathbf{b}_{i,j}$ are matrix transformation parameters learned on image graph nodes, $\mathbf{h}_i$ is the latent vector for $v_{i}$.

After the above sampling, we obtain the dynamic image relevant nodes ${v'}_{i,j}$ based on ${v}_{i}$. According to the Equation~\ref{messagecal}, we need to learn the affinity matrix $A$ and transformation matrix $\mathbf{w}$ for calculating the message. To effectively achieve depth-sensitive features, we generate hybrid depth-dependent filter weights and affinity matrices based on the multi-scale depth feature. Because of the ill depth prior, we do not sample the corresponding depth nodes, but generate another walk $\Delta\mathbf{\tilde{d}}_{i,j}$ for uniformly neighboring nodes on the depth feature graph.
% In particular, 
For stage $l$, we have obtained dynamic walked nodes ${v'}_{i,j}^l$ and $\mathit{\tilde{v}}_{i,j}^{l}$ from image and depth feature graph respectively. Note that we have aligned inconsistent scale depth features via down- or up-sampling operation. Then depth feature nodes $\mathit{\tilde{v}}_{i,j}^{l}$ learns to further generate affinity matrix $A_{i,j}^{l}$ and transformation matrix $\mathbf{w}_{i,j}^{l}$, as:
\begin{equation}
\begin{aligned}
	\mathbf{m}_i^{t+1} = \sum_{l \in L} \sum_{j \in \mathcal{N}(i)} \beta_l A_{i,j}^{l} \mathbf{h'}_j^{l,(t)} \mathbf{w}_{i,j}^{l}
	\label{messagecal_dep}
\end{aligned}
\end{equation}
\begin{equation}
\begin{aligned}
	\{A_{i,j}^{l} ;\mathbf{w}_{i,j}^{l}\} = \mathbf{\tilde{W}}_{i,j}^{l} \mathit{\tilde{v}}_{i,j}^{l} + \mathbf{\tilde{b}}_{i,j}^{l}
	\label{aff_w}
\end{aligned}
\end{equation}
In Equation~\ref{messagecal_dep}, $L$ stands for the layer from different level stages;
$\mathbf{h'}_j^{l,(t)}$ is the latent vector for dynamic nodes ${v'}_{i,j}^l$ from stage $l$ with node walk $\Delta\mathbf{d}_{i,j}$, and $\beta_l$ is the balance weight for depth feature maps integration from different stages.
$\mathbf{\tilde{W}}_{i,j}^{l}$ and $\mathbf{\tilde{b}}_{i,j}^{l}$ in Equation~\ref{aff_w} are matrix transformation parameters generated by depth nodes.
The calculated message is summarized as:
\begin{equation}
\begin{aligned}
	\mathbf{m}_i^{t+1} = \sum_{l \in L} \sum_{j \in \mathcal{N}(i)} \beta_l A_{i,j}^{l} \delta(\mathbf{h'}_j^{l,(t)}|\mathcal{V};j;\Delta\mathbf{d}_{i,j}) \mathbf{w}_{i,j}^{l}
	\label{messagecal_all}
\end{aligned}
\end{equation}
where  $\delta(\cdot)$ is a bilinear sampler which samples a new node  $\mathbf{h'}_j^{l,(t)}$ calculated by $\Delta\mathbf{d}_{i,j}$ over the whole nodes $\mathcal{V}$ of graph.

\subsection{Model instantiation}
The introduced architecture is shown in Figure~\ref{fig_arch} and a more detailed scheme of the DDMP module is further depicted in Figure~\ref{fig_dgmi}. 
DDMP module is embedded in the network within Stage II and III to propagate the context- and depth-aware message.
%To align different levels depth features dimension with image features,
To instantiate it, we first extract the hierarchical depth features from Stage II, III, and IV with different sizes, and adopt stride MaxPooling or interpolation operation to align these feature maps to the same size with corresponding stage image feature maps. 
Then we transform the channels of RGB and depth features from different stages to a fixed number C (\ie, 256) with $1\times1$ convolutions before feeding them into DDMP. 
DDMP accepts RGB features as input feature nodes $\mathbf{F}$, $\mathbf{F} \in R^{C \times H \times W}$, where $C$, $H$ and $W$ are the channel, height, and width of the feature map, respectively. We denote $\mathbf{H}^{(0)}$ as an initial state of the latent feature map $\mathbf{H}$, which has the same dimension with $\mathbf{F}$, and $\mathbf{H}^{(0)} = \mathbf{F}$. 
The dynamic walk $\Delta\mathbf{d}$ for each image node is generated by applying $3 \times 3$ convolutional layers according to Equation~\ref{deltad}. The same operation is done on depth features by generating $\Delta\mathbf{\tilde{d}}$ to obtain the sampled depth nodes.
Hybrid dynamic affinity matrices and weights in Equation \ref{aff_w} are also calculated on hierarchical depth features $\mathbf{F}_{dep}^{l} \in R^{C \times H \times W}$ by applying $3 \times 3$ convolutional layers, where $l$ stands for stage II / III / IV. Then we obtain affinity matrices $A^l \in R^{H \times W \times K}$ and another weights $W^l \in R^{H \times W \times K \times G}$, where $K$ is sampled size (\ie $3 \times 3$) and G is group size. All message $\mathbf{M} \in R^{C \times H \times W}$ is calculated as Equation \ref{messagecal_all} using group convolutional layers and concated with $\mathbf{F}$, to produce a refined feature map $\mathbf{H}^{(1)}$ via a $1\times1$ convolutional, which is described as Equation \ref{messageupd}. 
We perform $T=1$ to balance the performance and efficiency.

\subsection{Center-aware depth feature encoding (CDE)}
\label{cde}
Generally, the depth map estimated from off-the-shelf algorithms sometimes inevitably lose appearance details or fail to discriminate the depth between the foreground instance and backgrounds, 
which leads to unreliable depth prior for the depth-assisted 3D object detection.
It is already proved that multi-task strategy \cite{yang2018hdnet, He_2020_CVPR} can boost each single task to some degree, benefiting from the multi-fold regularization effect in the joint-optimization.
Hence, we augment an auxiliary task to jointly optimize with the main 3D detection task as depicted in Figure~ \ref{fig_arch}.
The augmented task with $xyz$ (Equation~\ref{loss_dep}) supervision in 3D space uniquely determines a point in 2D image plane, which imposes spatial constraints on the network to gain a better 3D instance-level understanding.
% towards the object to detect.
With the better instance-awareness brought by CDE, our model is able to alleviate the inaccurate depth prior in situation like \textit{occlusion} and \textit{distant} objects.
We adopt a similar network architecture for depth branch with the head only predicting 3D centers without predefined anchors. 
During training, we jointly optimize the losses of the two sub-branches in the light of guiding the intermediate depth features to be center-aware and thus enhancing the performance for 3D object detection. 
In this way, the proposed approach becomes more robust to the estimated depth accuracy, validation experiments can be found in Section \ref{sec_eval}.
The auxiliary task is omitted at the inference phase without extra computational cost.

\input{table/table_all_test}

\input{table/table_all_val}

\subsection{Objective functions}
We follow single-stage approaches \cite{roddick2018orthographic,ding2019learning} to pre-define anchors and then regress the positive samples. Firstly, 2D anchors $[\hat x,~\hat y,~\hat w,~\hat h]_{2d}$ are defined in 2D space for the 2D bounding box. 
$[\hat x,~\hat y]_{p}$ represents the 2D location projected from the 3D center.
3D anchor parameters $[\hat z, \hat w, \hat h, \hat l, \hat \theta]_{3d}$ denote 3D object center depth, physical dimension, and rotation, respectively.
Then all ground truth 3D boxes are projected to 2D space to compute the intersection over union (IoU) with 2D anchors. Positive 2D anchors with IoU $\ge 0.5$ combined with 3D anchor parameters are thus selected to participate in the regression for 3D bounding predictions. 
The predicted 2D-3D parameters include ($[x,y,w,h]_{2d}, [x,y]_{p}, [z,w,h,l,\theta]_{3d}$), which denote the classification scores, 2D bounding box locations, 3D center projections on 2D plane, 3D object center depths, physical dimensions and rotations, respectively. Similar to YOLOv3 \cite{redmon2018yolov3}, we adopt ($[t_x,t_y,t_w,t_h]_{2d}, [t_x,t_y]_{p}, [t_z,t_w,t_h,t_l,t_\theta]_{3d}$) to parameterize the corresponding predictions directly generated by the network. Hence the predicted 3D boxes is,
\begin{equation}
\begin{aligned}
	{[x,y]_{2d}} &  = [\hat x,\hat y]_{2d} + [t_x, t_y]_{2d}*[\hat w, \hat h]_{2d}\\
	{[w,h]_{2d}} & = [\hat w,\hat h]_{2d}*exp([t_w, t_h]_{2d})\\
	{[x,y]_{p}} & = [\hat x,\hat y]_{p} + [t_x, t_y]_{p}*[\hat w, \hat h]_{2d}\\
	{[w,h,l]_{3d}} & = [\hat w,\hat h, \hat l]_{3d}*exp([t_w, t_h, t_l]_{3d})\\
	{[z,\theta]_{3d}} & = [\hat z,\hat \theta]_{3d} + [t_z, t_\theta]_{3d},
\end{aligned}
\end{equation}
where we use the same anchor for $[\hat x,\hat y]_{2d}$ and $[\hat x,\hat y]_{p}$. Therefore, the loss for the detection branch $L_{det}$ contains classification loss, 2D bounding box regression loss and 3D box regression loss. 
We apply standard cross-entropy loss for classification and smooth L1 ($\tilde{L}1$) loss for box regression. The detection branch loss then is,
\begin{equation}
\begin{aligned}
	L_{det} & = L_{cls} + L_{2d} + L_{3d}\\
	%L_{cls} & = -log(s_t)\\
	L_{2d} & = \mathit{\tilde{L}1}([t_x,t_y,t_w,t_h]_{2d},\
        	[t_x^{gt},t_y^{gt},t_w^{gt},t_h^{gt}]_{2d})\\
	L_{3d} & = \mathit{\tilde{L}1}([t_x,t_y]_{p},[t_x^{gt},t_y^{gt}]_{p})\\
	& + \mathit{\tilde{L}1}([t_z,t_w,t_h,t_l,t_\theta]_{3d},\
	[t_z^{gt},t_w^{gt},t_h^{gt},t_l^{gt},t_\theta^{gt}]_{3d}),
	\label{loss_det}
\end{aligned}
\end{equation}
where $(*)^{gt}$ means (*)'s corresponding ground truth target.

For the auxiliary task of the depth branch, we regress the depth $[t'_z]_{3d}$ of the 3D object as well as the $[t'_x,t'_y]_{p}$ for projecting the 3D center onto the 2D image plane, which shares the same ground truth with the detection branch:
\begin{equation}
\begin{aligned}
	L_{dep}  = \mathit{\tilde{L}1}([t'_x,t'_y]_{p},[t_x^{gt},t_y^{gt}]_{p})
	+ \mathit{\tilde{L}1}([t'_z]_{3d},[t_z^{gt}]_{3d})
	\label{loss_dep}.
\end{aligned}
\end{equation}

The final training loss is a summation of $L_{det}$ and $L_{dep}$.
Inspired by focal loss, we utilize classification scores $s_t$ to balance the samples:
\begin{equation}
\begin{aligned}
	L  = (1-s_t)^\gamma(L_{det} + L_{dep}),
	\label{loss_all}
\end{aligned}
\end{equation}
where $\gamma$ is the focus parameter and set as 0.5.

\begin{figure*}[t]
 \centering
 \includegraphics[width=16.6cm]{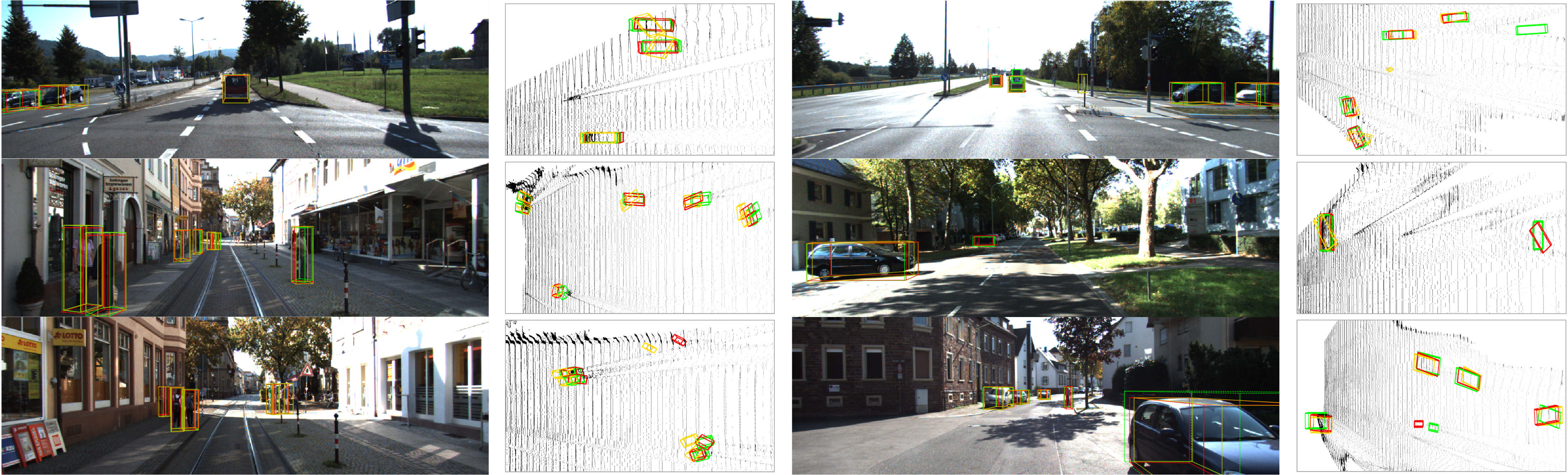}
 \caption{Qualitative comparison of ground truth (green), the baseline (yellow), and our method (red) on KITTI val set. For better visualization, the first and second columns show RGB and BEV images of point clouds converted from pre-estimated depth, respectively.}
 \label{fig_results}
\end{figure*}

\input{table/table_cyclist}

%% file: table/table_all_test.tex
\begin{table*}[h]
 \caption{Comparison with SoTA methods on the KITTI test set at IoU = 0.7. Our DDMP-3D achieves new SoTA performance.}
 \small
 \centering
 \begin{tabular}{|c|c|c|ccc|ccc|c|}
  \hline
  \multirow{2} * {Method} & \multirow{2} * {Reference} & \multirow{2} * {Speed (FPS)}& \multicolumn{3}{c|}{$\rm AP_{3D}$} & \multicolumn{3}{c|} {$\rm AP_{BEV}$}      &  \multirow{2} * {GPU}                \\
                       &   & & Mod.          & Easy          & Hard          & Mod.          & Easy          & Hard        &  \\
  \hline
  \hline
  
    FQNet\cite{liu2019deep}                  &CVPR 2019           & 2         & 1.51  & 2.77  & 1.01 &3.23&5.40&2.46 & 1080Ti\\
    ROI-10D\cite{manhardt2019roi}            &CVPR 2019           & 5         & 2.02  & 4.32  & 1.46 &4.91&9.78&3.74 & -\\
%    MonoGRNet\cite{qin2019monogrnet}        &AAAI 2019           & 16        & 5.74  & 9.61  & 4.25 &6.08&8.41&4.94 & Tesla P40\\
    MonoDIS\cite{simonelli2019disentangling} &ICCV 2019           & -         & 7.94  & 10.37 & 6.40 &13.19&17.23&11.12 & Tesla V100\\
    MonoPair\cite{chen2020monopair}          &CVPR 2020           & 17        & 9.99  & 13.04 & 8.65&14.83&19.28&12.89 &-\\
    UR3D \cite{shi2020distance}              &ECCV 2020           & 8         & 8.61  & 15.58 & 6.00  &12.51&21.85&9.2 &GTX Titan X\\
    RTM3D\cite{RTM3D}                        &ECCV 2020           & 20         & 10.34 & 14.41 & 8.77&14.20&19.17&11.99 &1080Ti\\
    AM3D\cite{ma2019accurate}                &ICCV 2019           & 3         & 10.74 & 16.50 & 9.52&17.32&25.30&14.91 &1080Ti\\
    % PatchNet\cite{Ma_2020_ECCV}            &ECCV 2020           & 3         & 11.12 & 15.68 & \textbf{10.17}&16.86&22.97&\textbf{14.97}\\
    DA-3Ddet\cite{Ye_2020_ECCV}              &ECCV 2020           & 3         & 11.50 & 16.77 & 8.93 & 15.90 & 23.35 & 12.11 & Titan RTX\\
    $\rm D^4LCN$\cite{ding2019learning}      &CVPR 2020           & 5         & 11.72 & 16.65 & 9.51&16.02&22.51&12.55 & 1080Ti\\ %Tesla V100\\
    Kinematic3D\cite{brazil2020kinematic}    &ECCV 2020           & 8         & 12.72 &19.07 & 9.17 & 17.52 & 26.69 & 13.10 & -\\
    \hline
    Ours                                     &-                   & 6         & \textbf{12.78} & \textbf{19.71} & \textbf{9.80}&\textbf{17.89}&\textbf{28.08}&\textbf{13.44} & Tesla V100 \\
  \hline
 \end{tabular}
 \label{tab_test}
\end{table*}

%% file: table/table_all_val.tex
\begin{table*}[h]
 \caption{Comparison on the KITTI ``val1'', ``val2'' set. We report the average precision (in \%) including $AP|_{R40}$ and $AP|_{R11}$ of ``Car'' on 3D object detection ($\rm AP_{3D}$) at IoU = 0.7.}
 \small
 \centering
  \begin{tabular}{|c|ccc|ccc|}
   \hline
   \multirow{2} * {Method} & \multicolumn{3}{c|}{Val1 [$AP|_{R40}$ / $AP|_{R11}$]}& \multicolumn{3}{c|}{Val2 [$AP|_{R40}$ / $AP|_{R11}$]}                          \\
                          & Mod.                     & Easy                      & Hard& Mod.                     & Easy                      & Hard \\
   \hline
   \hline
   OFT-Net\cite{roddick2018orthographic}          &\quad-\quad/ 3.27 &\quad-\quad/ 4.07&\quad-\quad/ 3.29        & \quad-\quad/\quad-\quad\quad&\quad-\quad/\quad-\quad\quad&\quad-\quad/\quad-\quad\quad   \\
   FQNet\cite{liu2019deep}                        &\quad-\quad/ 5.50 &\quad-\quad/ 5.98 &\quad-\quad/ 4.75        & \quad-\quad/ 5.11 &\quad-\quad/ 5.45&\quad-\quad/ 4.45\\
   ROI-10D\cite{manhardt2019roi}                  &\quad-\quad/ 6.93 &\quad-\quad/ 10.25 \quad&\quad-\quad/ 6.18        & \quad-\quad/\quad-\quad\quad&\quad-\quad/\quad-\quad\quad&\quad-\quad/\quad-\quad\quad \\
   MonoDIS\cite{simonelli2019disentangling}       &\quad-\quad/ 7.60 &\quad-\quad/ 11.06 &\quad-\quad/ 6.37        & \quad-\quad/\quad-\quad\quad&\quad-\quad/\quad-\quad\quad&\quad-\quad/\quad-\quad\quad\\
   MonoGRNet\cite{qin2019monogrnet}               & 7.56 / 10.19        & 11.90 / 13.88    & 5.76 / 7.62 & \quad-\quad/\quad-\quad\quad&\quad-\quad/\quad-\quad\quad&\quad-\quad/\quad-\quad\quad\\
   GS3D\cite{li2019gs3d}                          &\quad-\quad/ 10.97 &\quad-\quad/ 13.46 &\quad-\quad/ 10.38        & \quad-\quad / 10.51& \quad-\quad/ 11.63& \quad-\quad/ 10.51\\
   shift R-CNN\cite{naiden2019shift}              &\quad-\quad/ 11.29 &\quad-\quad/ 13.84 &\quad-\quad/ 11.08        &  \quad-\quad/\quad-\quad\quad                    &  \quad-\quad/\quad-\quad\quad    &  \quad-\quad/\quad-\quad\quad\\
   MonoPSR\cite{ku2019monocular}                  & \quad-\quad/ 11.48 & \quad-\quad/ 12.75 & \quad-\quad/ 8.59     &  \quad-\quad/ 12.24& \quad-\quad/ 13.94& \quad-\quad/ 10.77\\
   SS3D\cite{jorgensen2019monocular}              &\quad-\quad/ 13.15 &\quad-\quad/ 14.52 &\quad-\quad/ 11.85        &  \quad-\quad/ 8.42& \quad-\quad/ 9.45& \quad-\quad/ 7.34\\
   RTM3D\cite{RTM3D}                       &\quad-\quad/ 16.86 &\quad-\quad/ 20.77 &\quad-\quad/ 16.63         & \quad-\quad/ 16.29&\quad-\quad/ 19.47&\quad-\quad/ 15.57\\
   M3D-RPN\cite{brazil2019m3d}                    & 11.07 / 17.06                & 14.53 / 20.27              & 8.65 / 15.21                         & 10.07 / 16.48&14.51 / 20.40&7.51 / 13.34\\
   Pseudo-LiDAR\cite{wang2019pseudo}              & \quad-\quad/18.50 & \quad-\quad/ 28.20 & \quad-\quad/ 16.40      & \quad-\quad/\quad-\quad\quad&\quad-\quad/\quad-\quad\quad&\quad-\quad/\quad-\quad\quad\\
   Decoupled-3D\cite{cai2020monocular}            &\quad-\quad/ 18.68 &\quad-\quad/ 26.95 &\quad-\quad/ 15.82       &  \quad-\quad/\quad-\quad\quad& \quad-\quad/\quad-\quad\quad& \quad-\quad/\quad-\quad\quad\\
   UR3D \cite{shi2020distance}                    & 13.35 / 18.76             & 23.24 / 28.05              & 10.15 / 16.55                     &  11.10 / 16.75& 22.15 / 26.30& 9.15 / 13.60\\
   AM3D\cite{ma2019accurate}                      & \quad-\quad/ 21.09 & \quad-\quad/ \textbf{32.23}& \quad-\quad/ 17.26     & \quad-\quad/\quad-\quad\quad& \quad-\quad/\quad-\quad\quad& \quad-\quad/\quad-\quad\quad\\%& 10.74 & 16.50 & \textbf{9.52}  \\
   $\rm D^4LCN$\cite{ding2019learning}                   & 16.20 / 21.71     &  22.32 / 26.97     & 12.30 / 18.22            & \quad-\quad/ 19.54& \quad-\quad/ 24.29& \quad-\quad/ 16.38\\
   \hline
   Ours                                           & \textbf{20.39} / \textbf{23.12}  & \textbf{28.12} / 31.14  & \textbf{16.34} / \textbf{19.45}  &  \textbf{19.91} / \textbf{22.92} & \textbf{29.17} / \textbf{30.66} &  \textbf{15.26} / \textbf{18.75}\\
   \hline
  \end{tabular}
 \label{tab_all}
\end{table*}

%% file: table/table_cyclist.tex
\begin{table*}[h]
 \caption{3D object detection ($\rm AP_{3D}$) performance for ``Cyclist'' and ``Pedestrian'' on KITTI val split and test set (``val1''/test).}
 \small
 \centering
 \begin{tabular}{|c|ccc|ccc|}
  \hline
  \multirow{2} * {Method} & \multicolumn{3}{c|}{Cyclist} & \multicolumn{3}{c|} {Pedestrian}                             \\
           & Mod.               & Easy               & Hard               & Mod.              & Easy              & Hard             \\
  \hline
  \hline
  $\rm D^4LCN$\cite{ding2019learning}  & 4.41 / 1.67             & 5.85 / 2.45             & 4.14 / 1.36             & 11.23 / 3.42              & 12.95 / 4.55            & 11.05 / 2.83           \\
  Baseline                      & 4.07 / 0.17          & 4.50 / 0.32          & 4.08 / 0.17        & 6.93 / 2.32             & 8.11 / 3.05            & 6.78 / 1.81           \\
  Ours                          & \textbf{6.47 / 2.50} & \textbf{8.01 / 4.18} &  \textbf{6.27 / 2.32} & \textbf{12.11 / 3.55} & \textbf{14.42 / 4.93} & \textbf{12.05 / 3.01}\\
  \hline
 \end{tabular}
 \label{tab_cyclist}
\end{table*}

%% file: file/4-experiment.tex
\section{Experiments}
\label{sec_eval}
\paragraph{Dataset}
We conduct experiments on the KITTI dataset \cite{geiger2013vision,geiger2012we} of 7,481 and 7,518 images for training for testing, respectively. 
As in \cite{brazil2019m3d, ding2019learning}, we use two train-val splits of the KITTI dataset (``val1''/``val2'') to evaluate our DDMP-3D. 

\paragraph{Evaluation metrics}
Precision-recall curves are adopted for evaluation, and we report the average precision (AP) results of 3D and Bird’s eye view (BEV) object detection on KITTI validation and test set. 
The 40 recall positions-based metric $AP|_{R40}$ has been utilized by the KITTI test server instead of $AP|_{R11}$ since Aug. 2019. To compare with previous methods that only report $AP|_{R11}$ results, we also demonstrate $AP|_{R11}$ in all experiments except the comparison with state-of-the-art (SoTA) methods on KITTI test and validation set.
Three levels of difficulty are defined in the benchmark according to the 2D bounding box height, occlusion, and truncation degree, namely, ``Easy'', ``Mod.'', and ``Hard''. The KITTI benchmark ranks all methods based on the $\rm AP_{3D}$ of ``Mod.''.
As in~\cite{chen2017multi}, we adopt IoU = 0.7 as threshold for ``Car'' category. 
To validate the effectiveness on ``Cyclist'' and ``Pedestrian'' categories, we include the experiments with IoU = 0.5 for fair comparison. We denote AP for 3D and BEV as $\rm AP_{3D}$ and $\rm AP_{BEV}$, respectively.

\input{table/table_ablation}

\paragraph{Training details}
Initial anchors for detection heads are constructed as in $\rm D^4LCN$ \cite{ding2019learning}. For 2D anchors, 12 scales ranging from 30 to 400 pixels in height following the power function of $30*1.265^{n}, n = 0,...,11$, combining with aspect ratios of [0.5, 1.0, 1.5] to generate a total of 36 anchors. 
Mean statistics across matched 3D ground truth are also calculated as the initialization of 3D anchor parameters.
We employ ResNet-50 \cite{he2016deep} as feature extraction backbone for RGB and depth branch.
The input image size is scaled to $512\times1760$ and only horizontal flipping is applied to data augmentation. 2D space Non-Maximum Suppression (NMS) with an IoU threshold of 0.4 is finally used to drop the predicted bounding box redundancy. Note that the baseline method demonstrated in all experiments stands for simply dot-multiplication on RGB and depth features together, which is equivalent to treating each pixel on depth feature as a depth kernel.
The depth maps used for all experiments except for Table~\ref{tab_dep} are obtained by off-the-shelf monocular depth estimator DORN \cite{fu2018deep}.

Our model is trained by SGD with an initial learning rate 0.04, momentum 0.9, and weight decay 0.0005 respectively. We train the network with a batch size of 16 on 8 Nvidia Tesla v100 GPUs for 36k iterations.

\subsection{Comparison with state-of-the-arts}
\paragraph{Results on KITTI test and val set}
Table \ref{tab_test} and \ref{tab_all} show the 3D ``Car'' detection results on KITTI test and ``val1'', ``val2'' set at IoU = 0.7, respectively.
We report both $AP|_{R40}$ and $AP|_{R11}$ on validation set while only $AP|_{R40}$ on test set. In the KITTI leaderboard, our proposed DDMP-3D consistently outperforms other alternatives and ranks 1st compared with the top-ranked monocular-based 3D object detection methods.
Quantitatively, our method achieves the highest performance on ``Moderate'' set for both validation and test set, which is the main setting for ranking on the benchmark.
Large performance margins are observed over the second top-performed method ($\rm D^4LCN$ \cite{ding2019learning}), 4.19\% and 1.06\% on validation and test set in view of ``Mod.'', respectively. 
This phenomenon indicates that our outstanding performance benefits from the effective context- and depth-aware features learning via depth-conditioned dynamic graph message propagation and center-aware depth feature encoding.
%TODO need to modify
Note that some cutting-edge monocular methods use extra information for 3D object detection, \eg, AM3D \cite{ma2019accurate} designs two extra modules for background points segmentation and RGB information aggregation, while we attain the appealing results and acceptable inference speed without bells and whistles.

\paragraph{Results on ``Cyclist'' and ``Pedestrian''}
``Cyclist'' and ``Pedestrian'' categories are much more challenging than the ``Car'' for monocular 3D object detection due to their non-rigid structures and small scale.
We report these two categories respect to baseline and~\cite{ding2019learning} in Table~\ref{tab_cyclist}. 
Following \cite{ding2019learning}, $\rm AP_{3D}$ of ``Cyclist'' and ``Pedestrian'' on the ``val1''($AP|_{R11}$) and test($AP|_{R40}$) set at IoU = 0.5 are reported.
Thanks to our DDMP and CDE which provide better sensitivity to object location, we are able to localize these challenging categories to some degree and clearly outperform the alternatives.

\input{table/table_task}

\input{table/table_dep}

\subsection{Ablation study}
\paragraph{Main ablative analysis}
\label{ablation}
In Table \ref{tab_ablation}, we conduct ablation experiments to analyze the effectiveness of different components: 
I) Baseline: simply integrate depth features into RGB features by dot-product. 
II) DDMP (single-scale): depth-conditioned dynamic message propagation without multi-scale depth features. For the input of a certain DDMP, depth features from its corresponding single stage are extracted to generate affinity matrices and weights. For fair comparison, we keep the number of affinity matrices and weights the same with the following Group III.
III) DDMP (multi-scale): dynamic graph message propagation via hybrid and hierarchical depth features. For fair comparison, the parameter numbers are exactly kept the same as Group II (single-scale). The critical difference with respect to Group II lies in that the depth features fed into the DDMP are from different stages rather than a single stage.
IV) On the basis of Group III, CDE is added additionally to conduct a 3D center regression task with an auxiliary loss during the training phase, so as to guide the depth features to be center-aware.

As depicted in Table~\ref{tab_ablation}, we can observe that the performance continues to grow with the participation of components. Dynamic message propagation introduces an impressive increase from 18.82\% to 22.36\% on the moderate setting which confirms the effectiveness of the graph message mechanism between image and depth. Group III shows that hybrid affinity matrices and weights learning leads to 0.48\% performance gain on the moderate setting. It indicates that multi-scale depth features are useful in graph message propagation owing to the better depth perception on various scales of objects. Further, the auxiliary task brings a noticeable gain from 28.12\% to 31.14\% 
on the easy setting, which validates the effectiveness of the auxiliary task by learning center-aware depth features.

Qualitative comparisons of the baseline and our method are shown in Figure~\ref{fig_results}. The ground truth, baseline, and our method are colored in green, yellow, and red, respectively. For better visualization, the first and second columns show RGB images and BEV images of pseudo point clouds, respectively. Compared with the baseline, our DDMP-3D can produce higher-quality 3D bounding boxes in different kinds of scenes.
More quantitative and qualitative results are reported in our supplementary material.

\paragraph{Auxiliary task-guided depth encoding}
We explore the effect of different auxiliary tasks. We employ ``z'', ``xy'' and ``xyz'' center regressions on the depth branch (Equation~\ref{loss_dep}). As shown in Table~\ref{tab_task}, depth task ``z'' improves the moderate 3D detection performance from 22.84\% to 23.07\% while auxiliary center estimation task brings substantial improvements on all the three subsets from (22.84\% / 28.12\% / 19.09\%) to (23.13\% / 31.14\% / 19.45\%). The improvement is especially significant on ``Easy''. This further suggests that center-awareness is useful for object localization.

\paragraph{Impact of different depth estimators}
For generalization ability validation on different depth estimation methods, we choose the monocular depth estimator DORN \cite{fu2018deep} as well as the more accurate stereo matching method PSMNet \cite{chang2018pyramid} to obtain depth maps for comparison. 
As shown in Table~\ref{tab_dep}, the performance gain with respect to baseline and  $\rm D^4LCN$ \cite{ding2019learning} are enhanced with the increasing accuracy of estimated depth. Besides, improvements can be noticed both in monocular and stereo depth predictors.

%% file: table/table_ablation.tex
\begin{table*}[h]
 \caption{Ablative analysis on KITTI ``val1'' split set for $\rm AP_{3D}$ and $\rm AP_{BEV}$ at IoU = 0.7. Experiment group (I) is our baseline method. Different experiment settings are explored: (II) using depth-conditioned dynamic message propagation with single scale, (III) performing multi-scale dynamic message propagation, (IV) our full approach with auxiliary center-regression task.}
 \small
 \centering
 \begin{tabular}{|c|c|c|c|ccc|ccc|}
  \hline
  \multirow{2} * {Group} & DDMP & DDMP &\multirow{2} * {CDE} & \multicolumn{3}{c|}{$\rm AP_{3D}$} & \multicolumn{3}{c|} {$\rm AP_{BEV}$}                             \\
            & single-scale &  multi-scale          &       & Mod.          & Easy          & Hard          & Mod.          & Easy          & Hard                    \\
  \hline
  \hline
  I &          -&-           &-     & 18.82          & 26.03          & 16.27          & 24.18          & 33.06          & 19.63                \\
  II &\checkmark &-           &-     & 22.36          & 28.94          & 18.86          & 26.73          & 36.89          & 24.00                \\
  III &   -       & \checkmark &-     & 22.84          & 28.12          & 19.09          & 27.05          & 37.11          & 24.20                \\
  IV &- & \checkmark & \checkmark& \textbf{23.12} & \textbf{31.14} & \textbf{19.45} & \textbf{27.46} & \textbf{37.71} & \textbf{24.53}  \\
  \hline
 \end{tabular}
 \label{tab_ablation}
\end{table*}

%% file: table/table_task.tex
\begin{table}[h]
 \caption{Comparison of different auxiliary tasks on val split set.}
 \small
 \centering
 \begin{tabular}{|c|ccc|ccc|}
  \hline
  Center                            & \multicolumn{3}{c|}{$\rm AP_{3D}$} & \multicolumn{3}{c|} {$\rm AP_{BEV}$}                        \\
  Regression                        & Mod.          & Easy          & Hard          & Mod.          & Easy          & Hard             \\
  \hline
  \hline
    -                             & 22.84          & 28.12          & 19.09          & 27.05          & 37.11          & 24.20          \\
    only $z$                      & 23.07          & 28.11          & 19.19          & 26.95          & 36.70          & 24.06          \\
    $xy$                      & 22.51          & 28.68          & 18.52          & 26.80          & 37.15          & 21.38          \\
    $xyz$                         & \textbf{23.13} & \textbf{31.14} & \textbf{19.45} & \textbf{27.46} & \textbf{37.71} & \textbf{24.53} \\
  \hline
 \end{tabular}
 \label{tab_task}
\end{table}

%% file: table/table_dep.tex
\begin{table}[h]
 \caption{Comparison of different depth estimators on val split set at IoU = 0.7. The first, second, and third rows show the results of $\rm D^4LCN$ \cite{ding2019learning}, the baseline, and our method.}
 \small
 \centering
 \resizebox{86mm}{16mm}{
 \begin{tabular}{|c|c|ccc|}
  \hline
  \multirow{2} * {Depth} &\multirow{2} * {Method}& \multicolumn{3}{c|}{$\rm AP_{3D}$ / $\rm AP_{BEV}$}                                  \\
                                                          &          & Mod.          & Easy          & Hard                             \\
  \hline
  \hline
  \multirow{3} * {DORN\cite{fu2018deep}} 
                                        & $\rm D^4LCN$ & 21.71 / 19.54         & 26.97 / 24.29       & 18.22 / 16.38           \\
                                        & Baseline                    & 18.82 / 24.18         & 26.03 / 33.06       & 16.27 / 19.63          \\
                                        & Ours                        &\textbf{23.12} / \textbf{27.46} & \textbf{31.14} / \textbf{37.71} & \textbf{19.45} / \textbf{24.53} \\
  \hline
  \multirow{3} * {PSMNet\cite{chang2018pyramid}}                 
                                        & $\rm D^4LCN$      & 25.41 / \quad-\quad & 30.03 / \quad-\quad & 21.63 / \quad-\quad\\
                                        & Baseline       & 25.41 / 33.34         & 35.26 / 46.75       & 20.69 / 27.27           \\
                                        & Ours       & \textbf{30.83} / \textbf{36.20} & \textbf{41.76} / \textbf{52.87} & \textbf{24.78} /  \textbf{29.34} \\

  \hline
 \end{tabular}}
 \label{tab_dep}
\end{table}

%% file: file/5-conclusion.tex
\section{Conclusion}
We have presented a depth-conditioned dynamic message propagation (DDMP-3D) network, a novel graph-based approach that learns context- and depth-aware feature representation for 3D object detection. 
It dynamically samples context-aware nodes in the image context and predicts the hybrid filter weights and affinities based on the aligned multi-scale depth features for message propagation.
A center-aware depth encoding task is augmented and jointly trained with the whole network to resolve the challenge of inaccurate depth prior. 
This framework allows us to build a new state of the art among the monocular-based approaches, and this is demonstrated by the fact that we are rank $1^{st}$ in the highly competitive KITTI monocular 3D object detection track on the submission day (Nov 16th, 2020).

\section*{Acknowledgments}
This work was supported by 
by 
National Key R\&D Program of China (2019YFA0709502),
Shanghai Municipal Science and Technology Major Project (No.2018SHZDZX01), 
ZJLab, 
Shanghai Center for Brain Science and Brain-Inspired Technology, and
Shanghai Research and Innovation Functional Program (17DZ2260900).

%% file: file/6-appendix.tex
\appendix
\section{Architecture details}

As described in Section 3 in our paper, DDMP-3D contains image and depth feature encoding branches, with two DDMP modules adopted at Stage II and III, respectively.
We demonstrate the architecture details in Table \ref{tab_arch}. 
Since two DDMP modules (``DDMP\_1'' and ``DDMP\_2'') share a similar architecture, we only report the details of ``DDMP\_1''. ``DDMP\_1'' first integrates image features from stage II with depth features from stage II / III / IV (stage2\_depth2 / 3 / 4) and then concatenate the outputs together.
The outputs are scaled to the size of image features (``DDMP\_1 (update)'').
Note that the codes of constructing the model are attached in the supplementary material.

\input{table/table_arch_bak}

\section{Additional experiments}
\paragraph{Loss weight selection of auxiliary tasks.}
The loss weights for two branches in Equation 10 in our paper determine the influence of the auxiliary task on main task, which is a key hyper-parameter in our DDMP-3D framework.
To explore the sensitivity of this parameter, we conduct experiments as shown in Table~\ref{tab_task_weight}.

Paying more attention to the main task or at least equal weights to two tasks can achieve better performance for our model. When the weight for auxiliary task is equal to that of main task, it is favorable to detect objects on moderate and hard settings owing to its sensitivity to the centers. While it is friendly to detect easy objects with higher weight for main task.
A relatively high weight to auxiliary task brings slightly negative effect on the final performance.
This reflects that $L_{det}$ is essential on detection results whose weight should not be less than that of $L_{dep}$.

\paragraph{Statistic analysis on 3D metric.}
To further demonstrate the effectiveness of the proposed CDE, we compare the errors on the specific metrics (center ``xyz'') of the baseline method with or without CDE. 

\begin{figure*}[htb]
 \centering
 \includegraphics[width=17cm]{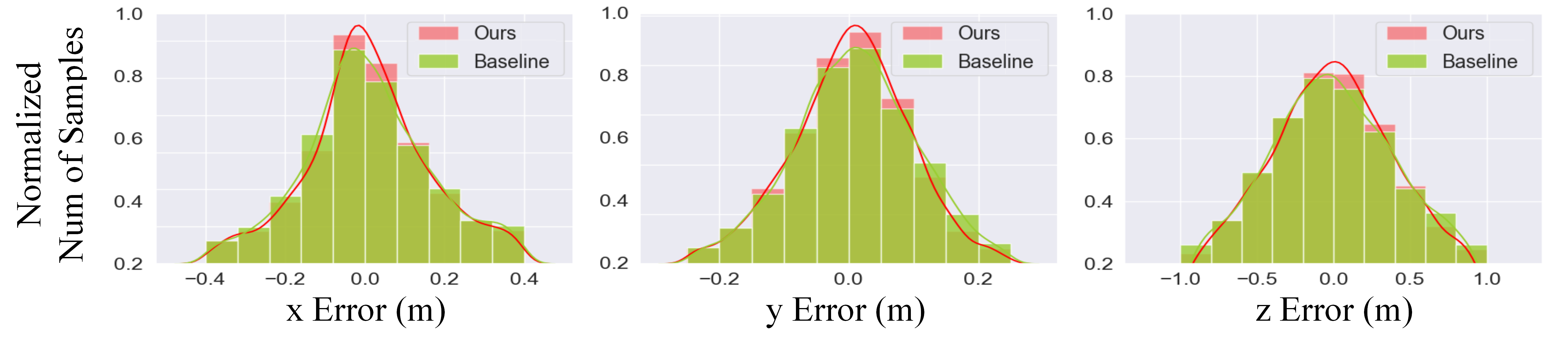}
 \caption{The statistic analysis and comparison of the baseline (green) and the baseline with our CDE (red). The vertical axis of the chart represents the number of samples after normalization. Improvements can be observed in the metrics ``x'', ``y'', and ``z''.}
 \label{fig_xyz}
\end{figure*}

As shown in Figure \ref{fig_xyz}, we can see that our proposed CDE improves the baseline method in ``x'', ``y'' and ``z'', resulting in more accurate monocular 3D object detection.
Note that the ``x'', ``y'', and ``z'' indicate the 3D camera coordinates of the object center point.

\paragraph{Ablation study on the auxiliary task for the depth encoding.}

We report the experiment results of deploying other auxiliary tasks in Table \ref{tab_task_more}.
It is observed that various auxiliary tasks have certain effects on the performances.
The task of 3D center regression (``xyz'') is critical, which introduces notable improvements on all settings.

However, the performance of adding 3D bounding box regression (``whl + rotation'') or classification task experienced a drop at some settings.
We consider that it is difficult for bounding box regression and classification on depth map without well-defined boundary and distinctive appearance.
Therefore, we further validate the hypothesis that the center-aware depth feature encoding helps monocular 3D object detection.

\paragraph{Different message propagation strategies.} 
How to effectively deliver the depth information through image feature domain and learn context- and depth-aware feature representation is critical for monocular 3D detection.
This is also the objective of this paper.
We perform different message propagation strategies to verify the effectiveness of our proposed DDMP-3D.

As shown in Table \ref{tab_gmp}, ``3DNet'' is the baseline in $\rm D^4LCN$ \cite{ding2020learning}, which only contains the single detection branch without the guidance of depth map.
``3DNet w/ DGMN~\cite{zhang2020dynamic}'' augments detection branch with the DGMN formulation to perform the effective feature learning in the RGB feature domain.
``Baseline'' integrates images with depth maps via a common multiplication operation. 
With the guidance of depth maps, it easily outperforms the above two methods. 
``3DNet w/ DGMN + Depth'' introduces the depth information via the common multiplication operation.
``DDMP'' is our proposed module for integrating image and depth via graph message propagation.

The large gains on all settings demonstrate its effectiveness on propagating depth-conditioned messages. 
Different with DGMN~\cite{zhang2020dynamic}, our proposed DDMP generates hybrid filters and affinities used for propagating message from the multi-scale sampled depth feature;
``DDMP + CDE'' augments the ``CDE'' task which has been discussed in the main paper. 
Note that thanks to the non-linear Softmax operation on the generated affinity matrix, the network learns from the normalized affinities to further boost the final detection performance, as is shown in the last two rows in Table \ref{tab_gmp}.

\begin{table*}[htb]
 \caption{Comparison results (3D ``Car'' detection) of different weights of auxiliary tasks on val split set (IoU = 0.7). $\alpha$ and $\beta$ are the weights for $L_{det}$ and $L_{dep}$, respectively.}
 \small
 \centering
  \resizebox{90mm}{15mm}{
 \begin{tabular}{|c|ccc|ccc|}
  \hline
  \multirow{2} * {$\alpha : \beta$}             & \multicolumn{3}{c|}{$\rm AP_{3D}$} & \multicolumn{3}{c|} {$\rm AP_{BEV}$}\\
                                    & Mod.           & Easy           & Hard           & Mod.           & Easy           & Hard     \\
  \hline
  \hline
    1:0                        & 22.84          & 28.12          & 19.09          & 27.05          & 37.11          & 24.20          \\
    1:2                        & 22.71          & 31.35          & 18.94          & 27.18          & 37.96          & 24.38          \\
    2:1                        & 22.85          & \textbf{32.32} & 19.35          & 27.36          & \textbf{41.65} & 24.47          \\
    1:1                 & \textbf{23.13} & 31.14          & \textbf{19.45} & \textbf{27.46} & 37.71          & \textbf{24.53} \\
  \hline
 \end{tabular}}
 \label{tab_task_weight}
\end{table*}

\begin{table*}[t]
 \caption{Comparison results (3D ``Car'' detection) of different auxiliary tasks on val split set (IoU = 0.7). ``DDMP + bbox'', ``DDMP + class'', and ``DDMP + center'' stand for the bounding boxes regression, classification and center regression tasks, respectively.}
 \small
 \centering
 \resizebox{178mm}{16mm}{
 \begin{tabular}{|c|ccc|ccc|}
  \hline
  \multirow{2} * {Method}             & \multicolumn{3}{c|}{$\rm AP_{3D}$} & \multicolumn{3}{c|} {$\rm AP_{BEV}$}                          \\
                                    & Mod.           & Easy           & Hard           & Mod.           & Easy           & Hard          \\
  \hline
  \hline
    DDMP                        & 22.84          & 28.12          & 19.09          & 27.05          & 37.11          & 24.20          \\
    DDMP + bbox           & 22.48 \textcolor{green}{(-0.36)} & 28.84 \textcolor{red}{(+0.72)} & 18.31 \textcolor{green}{(-0.78)} & 26.06 \textcolor{green}{(-0.99)} & 36.35 \textcolor{green}{(-0.76)} & 21.00 \textcolor{green}{(-3.20)} \\
    DDMP + class           & 22.69 \textcolor{green}{(-0.15)} & 28.72 \textcolor{red}{(+0.60)} & 19.16 \textcolor{red}{(+0.07)} & 26.94 \textcolor{green}{(-0.09)} & 36.87 \textcolor{green}{(-0.24)} & 24.11 \textcolor{green}{(-0.09)} \\
    DDMP + center          & \textbf{23.13} \textcolor{red}{(+0.29)} & \textbf{31.14} \textcolor{red}{(+3.02)} & \textbf{19.45} \textcolor{red}{(+0.36)} & \textbf{27.46} \textcolor{red}{(+0.41)} & \textbf{37.71} \textcolor{red}{(+0.60)} & \textbf{24.53} \textcolor{red}{(+0.33)} \\
  \hline
 \end{tabular}}
 \label{tab_task_more}
\end{table*}

\begin{table*}[htb]
 \caption{Comparison results (3D ``Car'' detection) of different message integration positions on val split set (IoU = 0.7).}
 \small
 \centering
 \resizebox{178mm}{23mm}{
 \begin{tabular}{|c|c|c|ccc|ccc|}
  \hline
  \multirow{2} * {Method}  & \multirow{2}*{Image input} &  \multirow{2}*{Depth map input}           & \multicolumn{3}{c|}{$\rm AP_{3D}$} & \multicolumn{3}{c|} {$\rm AP_{BEV}$}                          \\
                                  &&  & Mod.           & Easy           & Hard           & Mod.           & Easy           & Hard          \\
  \hline
  \hline
    3DNet     & \multirow{2} * {\checkmark} & \multirow{2} * {-}                   & 14.61          & 17.94          & 12.74          & 19.89          & 24.87          & 16.14          \\
    3DNet w/ DGMN~\cite{zhang2020dynamic}  &  &             & 16.98          & 20.12          & 15.17          & 21.49          & 26.40          & 17.96          \\
    \hline
    Baseline (3DNet + Depth)     & \multirow{2} * {\checkmark} & \multirow{2} * {\checkmark} & 18.82          & 26.03          & 16.27          & 24.18          & 33.06          & 19.63                \\
    3DNet w/ DGMN~\cite{zhang2020dynamic} +Depth  &  &          & 19.59          & 27.78          & 16.48          & 25.30          & 35.59          & 20.32          \\
  \hline
    DDMP    & \multirow{3} * {\checkmark} & \multirow{3} * {\checkmark}         & 22.84          & 28.12          & 19.09          & 27.05          & 37.11          & 24.20          \\
    DDMP + CDE &  &   & 23.13 & 31.14 & \textbf{19.45} & 27.46 & 37.71 & 24.53 \\
    DDMP (\textit{Softmax}) + CDE  &  &         & \textbf{23.17} & \textbf{32.40} & 19.35 & \textbf{27.85} & \textbf{42.05} & \textbf{24.91} \\
  \hline
 \end{tabular}}
 \label{tab_gmp}
\end{table*}

\begin{figure*}[htb]
 \begin{center}
  \includegraphics[width=17.5cm, height=2cm]{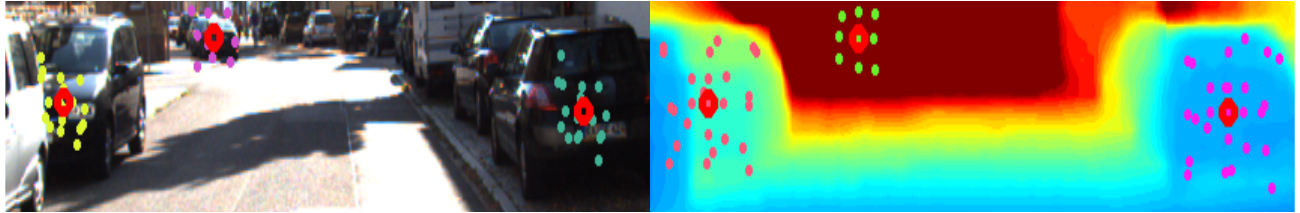}
 \end{center}
 \caption{Visualization of sampling points on the images and depth maps, and the predicted results on the KITTI dataset.}
 \label{fig1}
\end{figure*}

\begin{figure*}[htp]
 \begin{center}
  \includegraphics[width=17.8 cm]{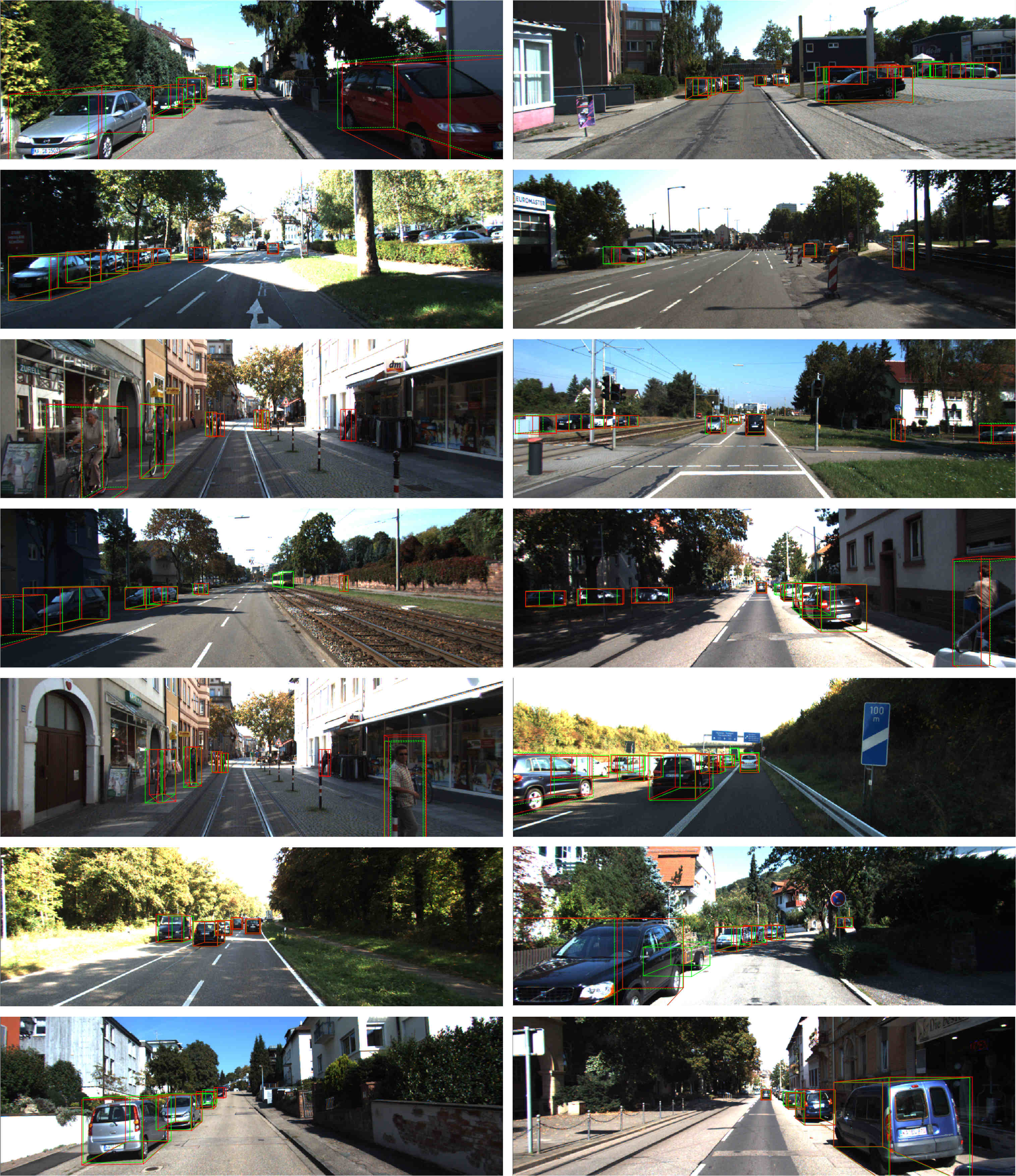}
 \end{center}
 \caption{More qualitative results on the KITTI dataset. 
 The 3D ground-truth boxes and our DDMP-3D predictions are drawn in green and red, respectively.}
 \label{fig2b}
\end{figure*}

% \clearpage

\section{Additional qualitative results}
\paragraph{Visualization of dynamic sampling points.}
Figure \ref{fig1} shows dynamic sampling points based on the learned $\Delta d$ and $\Delta \Tilde{d}$ from images and depth maps, respectively.
The receiving nodes are shown with red circles.
As shown in left figure, our sampled image nodes accurately perceive the semantic context: object boundary of left car and the small object, to enable more effective message passing.
Also, in right figure, we demonstrate our dynamically sampled multi-scale depth nodes that dedicate to capture the context of the target objects.

\paragraph{More qualitative results.}
Figure \ref{fig2b} shows more qualitative results on the KITTI dataset. 
The 3D ground-truth boxes and our DDMP-3D predictions are drawn in green and red, respectively.
As clearly observed, DDMP-3D can produce high-quality 3D bounding boxes in various scenes.

%% file: table/table_arch_bak.tex
\begin{table*}[h]
 \caption{Detailed architecture. The table expands the details of the DDMP process on image stage II (DDMP\_1), including the message propagation from image stage II and depth stage II / III / IV, and the message updating on image stage II.}
 \small
 \centering
 \begin{tabular}{|c|c|c|c|}
  \hline
  Module & Type / Stride &    Input name &  Output name: size                         \\
  \hline
  \hline
  \multirow{5}*{Detection backbone (ResNet-50)} 
  &   conv1 / s=2 &   image 
  &  img\_conv1:  $64 \times 256 \times 880$ \\
  \cline{2-4}
  &   conv2\_x / s=2 &  img\_conv1
  &  img\_stage1:  $256 \times 128 \times 440$ \\
  \cline{2-4}
  &   conv3\_x / s=2 &  img\_stage1
  &  img\_stage2:  $512 \times 64 \times 220$ \\
  \cline{2-4}
  &   conv4\_x / s=2 &  img\_stage2
  &  img\_stage3:  $1024 \times 32 \times 110$ \\
  \cline{2-4}
  &   conv5\_x (dilated=2) / s=1&  img\_stage3
   &  img\_stage4:  $2048 \times 32 \times 110$ \\
  \hline
  \hline
  \multirow{5}*{Depth backbone (ResNet-50)} 
  &   conv1 / s=2 &  estimated depth map
  &  dep\_conv1:  $64 \times 256 \times 880$ \\
  \cline{2-4}
  &   conv2\_x / s=2 &  dep\_conv1
  &  dep\_stage1:  $256 \times 128 \times 440$ \\
  \cline{2-4}
  &   conv3\_x / s=2 &  dep\_stage1
  &  dep\_stage2:  $512 \times 64 \times 220$ \\
  \cline{2-4}
  &   conv4\_x / s=2 &  dep\_stage2
  &  dep\_stage3:  $1024 \times 32 \times 110$ \\
  \cline{2-4}
  &   conv5\_x (dilated=2) / s=1&  dep\_stage3
   &  dep\_stage4:  $2048 \times 32 \times 110$ \\
  \hline
  \hline
  \multirow{10}*{DDMP\_1 (stage2\_depth2)}
  &   conv $1 \times 1$ &  img\_stage2 
  &   img\_stage22: $256 \times 64 \times 220$ \\
  \cline{2-4}
  &   conv $3 \times 3$ &  img\_stage22 
  &   img\_stage22\_offset: $18 \times 64 \times 220$ \\
  \cline{2-4}
  &   deform\_unfold $3 \times 3$ &  img\_stage22 
  &   img\_stage22\_sample: $256 \times 9 \times 64 \times 220$ \\
  \cline{2-4}
  &   conv $1 \times 1$ &  dep\_stage2 
  &   dep\_stage22: $256 \times 64 \times 220$ \\
  \cline{2-4}
  &   deform\_conv $3 \times 3$ (group=1) &  dep\_stage22 
  &   dep\_stage22\_affinity: $9 \times 64 \times 220$ \\
  \cline{2-4}
  &   conv $3 \times 3$ &  dep\_stage22 
  &   dep\_stage22\_filter: $9 \times 64 \times 220$ \\
  \cline{2-4}
  &   \multirow{2}*{dot} &  img\_stage22\_sample;& \multirow{2}*{stage22\_sample: $256 \times 9 \times (64 * 220)$ }\\
  && dep\_stage22\_filter &   \\
  \cline{2-4}
  &   \multirow{2}*{matmul(group=1)} &  stage22\_sample;& \multirow{2}*{message\_stage22: $256 \times 64 \times 220$ }\\
  && dep\_stage22\_affinity &   \\
  
  \hline
  \multirow{2}*{DDMP\_1 (stage2\_depth3)}&   
  Similar to stage2\_depth2 &  img\_stage2;& \multirow{2}*{message\_stage23: $256 \times 64 \times 220$ }\\
  &(+ interpolate on dep\_stage3)& dep\_stage3 &   \\
  \hline
  \multirow{2}*{DDMP\_1 (stage2\_depth4)}&   
  Similar to stage2\_depth2 &  img\_stage2;& \multirow{2}*{message\_stage24: $256 \times 64 \times 220$ }\\
  &(+ interpolate on dep\_stage4)& dep\_stage4 &   \\
  \hline
  \multirow{2}*{DDMP\_1 (update)}&   
  \multirow{2}*{concat \& conv $3 \times 3$} &  img\_stage2;& \multirow{2}*{img\_stage2: $512 \times 64 \times 220$ }\\
  && message\_stage22/23/24 &   \\
  \hline
  \multirow{2}*{DDMP\_2}&   
  \multirow{2}*{Similar to DDMP\_1} &  img\_stage3;& \multirow{2}*{img\_stage3: $1024 \times 32 \times 110$ }\\
  && dep\_stage2/3/4 &   \\
  \hline
 \end{tabular}
 \label{tab_arch}
\end{table*}